\pdfoutput=1
\documentclass[11pt]{article}

\usepackage[final]{acl}

\usepackage{times}
\usepackage{latexsym}

\usepackage[T1]{fontenc}
\usepackage[utf8]{inputenc}

\usepackage{microtype}

\usepackage{inconsolata}

\usepackage{graphicx}
\usepackage{algorithm}
\usepackage{algpseudocode}
\usepackage{xcolor} 
\definecolor{iccvblue}{rgb}{0.21,0.49,0.74}
\usepackage{hyperref}
\usepackage{tcolorbox}
\usepackage{booktabs}
\usepackage{multirow}
\usepackage{colortbl}
\usepackage{amsmath}
\usepackage{xspace}
\usepackage{pifont}% http://ctan.org/pkg/pifont
\usepackage{color}
\definecolor{lightgreen}{RGB}{220, 255, 220}  % Very light green
\definecolor{lightblue}{RGB}{220, 235, 255}   % Very light blue

\newcommand{\lvlm}{\textsc{l-vlm}}
\newcommand{\svlm}{\textsc{s-vlm}}
\title{When Big Models Train Small Ones: Label-Free Model Parity Alignment for Efficient Visual Question Answering using Small VLMs}

\author{Abhirama Subramanyam Penamakuri\thanks{Equal contribution.}, Navlika Singh\footnotemark[1], Piyush Arora\footnotemark[1] and Anand Mishra \\
  Indian Institute of Technology Jodhpur\\
  \texttt{\{penamakuri.1,singh.119,arora.8,mishra\}@iitj.ac.in}\\
  \href{https://github.com/vl2g/MPA}{\textbf{{https://github.com/vl2g/MPA}}}}

\begin{document}
\maketitle
\begin{abstract}
Large Vision-Language Models (\lvlm{}s) have demonstrated remarkable performance in various vision and language tasks, including visual question answering (VQA). However, their high computational cost makes them impractical for resource-constrained settings and inference-heavy applications. In contrast, Small Vision-Language Models (\svlm{}s) offer efficiency but suffer from a significant performance gap compared to their larger counterparts. In this work, we introduce the Model Parity Aligner (MPA), a novel framework designed to systematically improve \svlm{}s by leveraging unlabeled images and effective knowledge transfer from \lvlm{}s. Instead of traditional knowledge distillation methods that rely on labeled training data, MPA employs a strategic parity-based approach that precisely identifies the knowledge disparities between \svlm{}s and \lvlm{}s, and optimizes training by targeting only these disparities. We conduct extensive experiments on four diverse VQA benchmarks, namely TextVQA, ST-VQA, ChartQA, and OKVQA, each of which required specialized reasoning capabilities such as text recognition, chart interpretation, and commonsense and factual understanding. Our results demonstrate that MPA consistently enhances the performance of \svlm{} on all benchmarks, reducing the performance gap while maintaining computational efficiency. We make our code publicly available.
\end{abstract}

% \section{Introduction}

\section{Introduction}
\begin{figure}[!t]
\centering
  \includegraphics[width=\columnwidth]{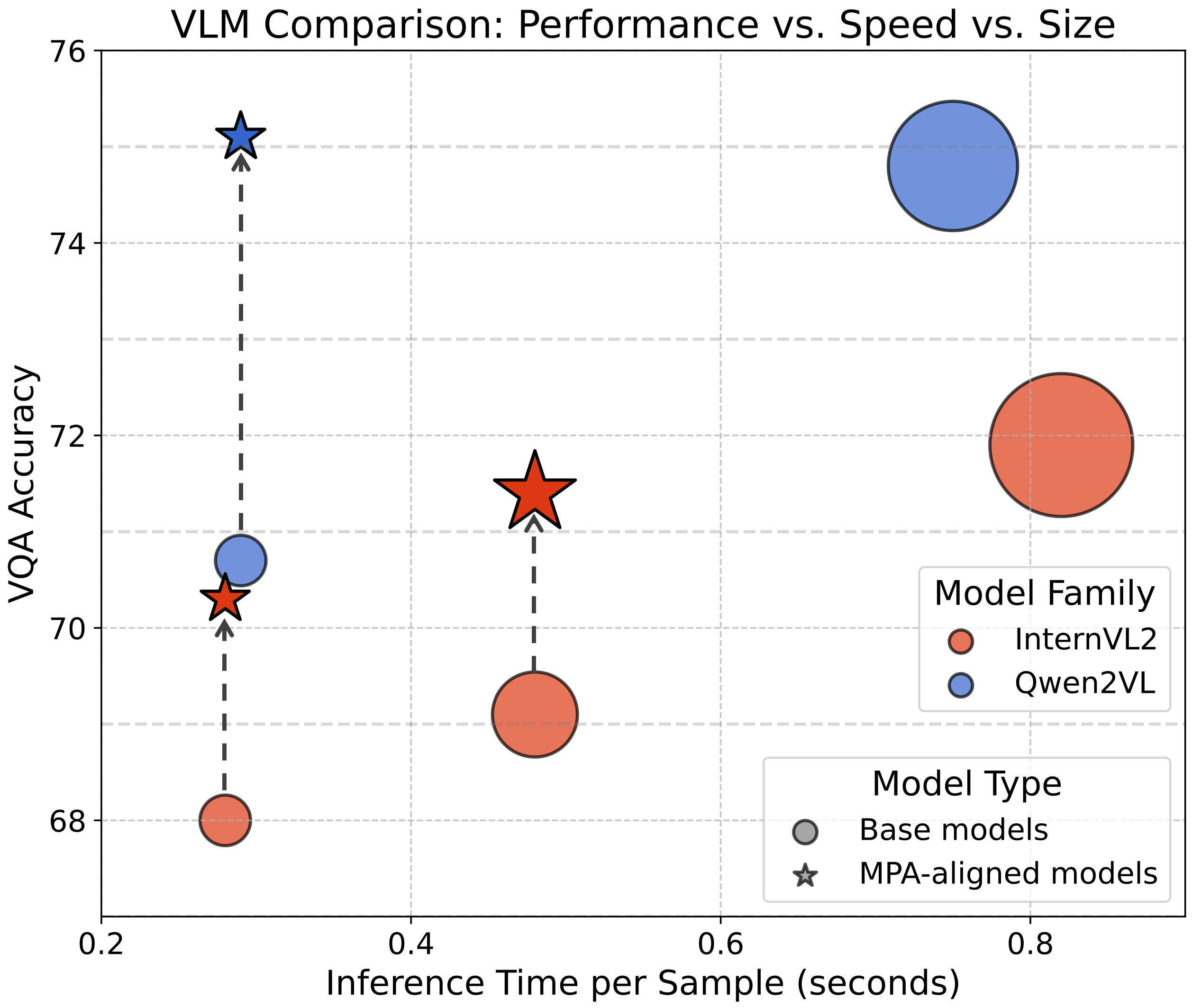}
  \caption{Small models often struggle to match the performance of their larger counterparts. We show model sizes using circle with radius proportional to the parameter count, and their respective inference time and VQA accuracies in X and Y-axis, respectively on one of the datasets used in this paper~\cite{textvqa}. Proposed MPA significantly enhances VQA accuracy for five \svlm{}s across four datasets. \textbf{(Best viewed in color)}.
  }
  \label{fig:goal}
\end{figure}

Large vision and language models (\lvlm{}s) have recently made remarkable progress on various vision and language tasks, including visual question answering (VQA)~\cite{liu2023llava,dai2024instructblip,li2023blip2,zhu2023minigpt,ye2023mplug,qwen2vl,chen2024internvl,ghosh2024exploring}. This makes them a de facto first choice for the VQA task on a new data set that does not have labeled training samples. However, \lvlm{}s may not be the most practical choice in resource-constrained settings and especially for inference-heavy tasks such as VQA, due to their high computational requirements and latency. In contrast, smaller vision and language models (\svlm{}s) are more efficient but fall significantly short in performance, as shown in Figure~\ref{fig:goal}. This raises a critical question: \emph{Can we improve \svlm{}s by a relevant and effective knowledge transfer from \lvlm{}s?} 

Several techniques have been explored to transfer knowledge from large neural models to smaller ones such as: (i) knowledge distillation (KD)~\cite{hinton2015distilling,sanh2019distilbert,guminillm,ko2024distillm,xu2024llavadi,shu2024llavamod,llavakd} trains a small model (student) to mimic a large model (teacher) by learning from its soft labels or intermediate representations. However, KD typically relies on labeled training data, which may not always be available, and effectively distilling multimodal knowledge remains a challenge due to the complex interplay between vision and language features. (ii) Adapter-based methods~\cite{houlsby2019parameter,hu2022lora,liu2022few,dettmers2023qlora} introduce lightweight trainable layers into large models to enable efficient fine-tuning. Although these methods reduce training costs, they still require access to large models during inference, limiting their practical advantages in resource-constrained environments. 
(iii) Self-supervised learning and pseudo-labeling~\cite{chen2013neil,veit2017learning,radosavovic2018data,xie2020self,seltda} provide an alternative by leveraging unlabeled data to generate training signals. However, naïve pseudo-labeling often propagates noisy predictions, reducing overall effectiveness. Moreover, the challenge of systematically transferring knowledge from large to small vision-language models using pseudo-labeling remains largely under explored. Addressing this gap is crucial for making smaller models more capable without the high computational cost of large models for inference.

To fill the aforementioned gaps, we introduce the \underline{M}odel \underline{P}arity \underline{A}ligner (MPA) -- a framework that enables effective knowledge transfer from \lvlm{} to \svlm{} using only unlabeled images. Instead of relying on traditional knowledge distillation or fine-tuning, MPA utilizes large model-guided pseudo-labeling with quality assessment. MPA  accurately identifies and addresses the knowledge gaps between \svlm{} and \lvlm{}, ensuring that small models learn from high-confidence predictions while minimizing error propagation. By leveraging the strong reasoning capabilities of large VLMs to create high-quality supervision signals through systematic parity assessment, MPA efficiently addresses performance gaps while maintaining computational efficiency. 

We conducted extensive experiments and ablation studies to evaluate the effectiveness of the MPA. Specifically, we used four public datasets -- TextVQA~\cite{textvqa}, ST-VQA~\cite{stvqa}, ChartQA~\cite{masry2022chartqa}, and OKVQA~\cite{okvqa}. These datasets require additional capabilities such as visual text understanding, chart interpretation, and world knowledge integration, making them well-suited to test the robustness of MPA. We experimented with ten combinations of \lvlm{} and \svlm{} pairs, demonstrating that MPA consistently improves \svlm{} performance across all benchmarks, highlighting its effectiveness in knowledge transfer.

\noindent\textbf{Contributions:}
(i) We propose a Model Parity Aligner (MPA) -- an effective approach that empowers small VLMs and improve their visual question-answering performance using only unlabeled images, eliminating the need for expensive labeled datasets. 
(ii) MPA employs a novel parity-based training paradigm, leveraging the \lvlm{} to generate pseudo-labels for unlabeled images while identifying and targeting specific knowledge gaps between \svlm{} and \lvlm{}. This strategy ensures reliable supervision, minimizes noise, and maximizes relevant knowledge transfer.
(iii) MPA achieves consistent improvement across four diverse VQA benchmarks. Furthermore, our findings indicate that MPA not only improves VQA performance, but also enables \svlm{} to benefit from closed-source \lvlm{}s and enhances its core capabilities beyond VQA, such as text recognition and text-aware captioning.

\section{Related Work}
\label{sec:related_work}
\textbf{Small and Large VLMs:} Following the success of large language models (LLMs)~\cite{devlin-etal-2019-bert,gpt3,touvron2023llama,falcon,dubey2024llama,yang2024qwen2technicalreport} across NLP tasks, vision and language models (VLMs)~\cite{liu2023llava,dai2024instructblip,zhu2023minigpt,ye2023mplug,chen2024internvl,qwen2vl,zhou2024tinyllava,smolvlm} have been developed that process both visual and textual data. 
Although state-of-the-art VLMs achieve impressive zero-shot performance, their growing parameter count impose significant constraints on computational efficiency, accessibility, and deployment costs. This trade-off between efficiency and capacity requires the development of smaller VLMs~\cite{zhou2024tinyllava,shao2024imp,smolvlm} that maintain competitive performance with reduced computational demands~\cite{lu2024small}. The key approach to developing \svlm{}s from \lvlm{}s involves substituting the internal LLM with lightweight alternatives~\cite{team2024gemma,abdin2024phi,zhang2024tinyllama,huggingfacetb2023smollm}. Inspired by the literature on LLM~\cite{lu2024small}, we follow a parameter-based taxonomy where VLMs with $\leq$ 5B parameters are classified as \svlm{}s, while those that exceed this threshold are \lvlm{}s. For context, a small 4B-parameter VLM constitutes just 0.2\% of the estimated 1.8T parameters of GPT-4.

\noindent\textbf{Knowledge Distillation:} It transfers knowledge from large teacher models to smaller student models using KL-divergence over soft logits~\cite{hinton2015distilling} or feature representations~\cite{dkmf,fnkd,sanh2019distilbert}. With LLMs adhering to scaling laws, their distillation has gained significant interest. Recent methods for LLMs~\cite{guminillm,ko2024distillm} and \lvlm{}s~\cite{shu2024llavamod,xu2024llavadi,llavakd} explore KL-Divergence variants, while others~\cite{hsieh2023distilling,tian2024beyond,ranaldi2024aligning} distill reasoning via LLM-generated Chain-of-Thought rationales. In contrast to standard KD, which distills over the labeled dataset, our method identifies and supervises only the samples that represent knowledge gaps between the student and teacher. This targeted strategy enables efficient, model-agnostic training using only input-output access to the teacher--including closed-source \lvlm{}s. 

\begin{figure*}[!t]
\centering
  \includegraphics[width=\textwidth]{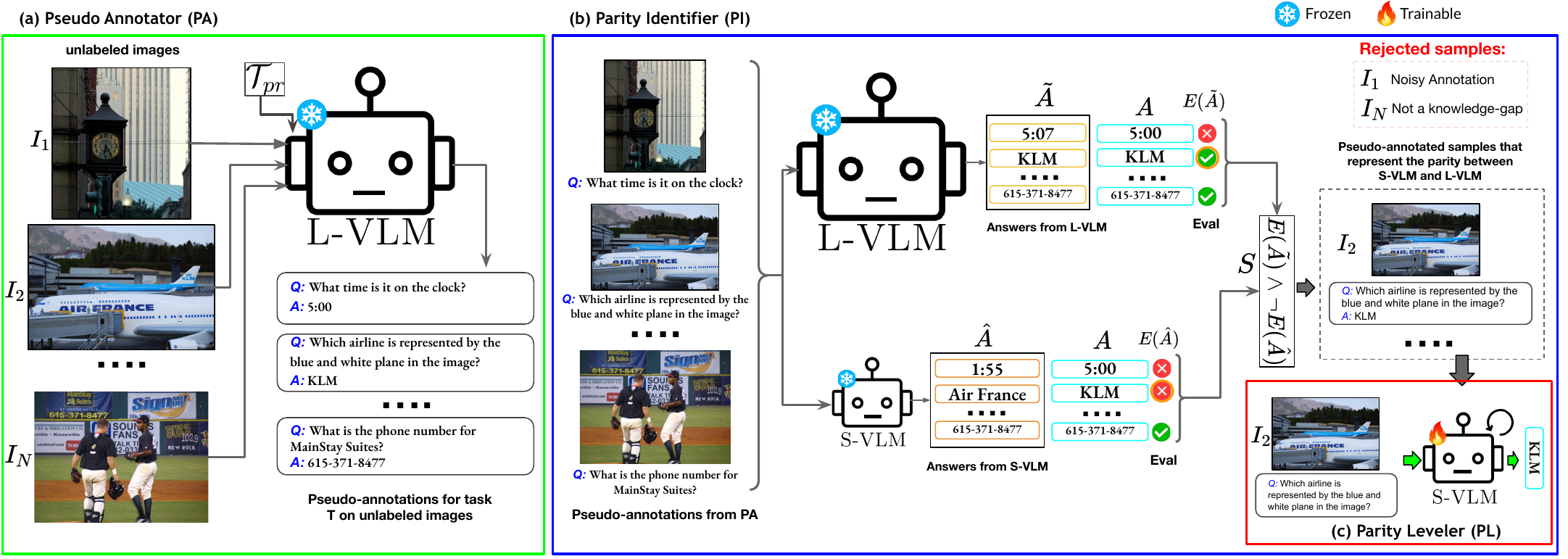}
  \caption{\textbf{Overview of the proposed MPA framework.} It consists of three modules, namely (a) Pseudo Annotator (Section~\ref{subsec:pa}), (b) Parity Identifier (Section~\ref{subsec:pi}), and (c) Parity Leveler (Section~\ref{subsec:pl}). Given a set of unlabeled images $\mathcal{I}$ and task $\mathcal{T}$, MPA begins with automatically annotating the unlabeled images, followed by strategic data selection that targets knowledge gaps of \svlm{} with the \lvlm{}, while accounting for annotation quality. This selection process identifies parity, capturing instances where the \lvlm{} answers correctly while the \svlm{} fails. Finally, PL updates the \svlm{}'s parameters on the obtained parity subset. \textbf{(Best viewed in color)}.}
  \label{fig:method}
\end{figure*}
\noindent\textbf{Data Augmentation for VQA:} 
Vision and language tasks such as VQA have traditionally been benefited by data augmentation, and visual question generation becomes a natural choice to generate augmented data~\cite{fan2018question,jain2017creativity,krishna2019information,mostafazadeh-etal-2016-generating,tag,jahagirdar2021look,groundedVQG,vedd2022guiding}. Although few methods~\cite{kddaug,concat,kil2021discovering,seltda} augment the data-scarce VQA datasets to improve performance, other methods~\cite{banerjee2021weaqa,changpinyo2022all} leverage large-scale image-caption datasets to generate noisy VQA labels and use them as VQA foundational data. Distinctively different from these lines of work, we employ \lvlm{}s to pseudo-label unlabeled images with a quality check to discard noise, ensuring minimal yet effective annotations for targeted improvements of \svlm{}s.

% which is a dual task~\cite{li2018visual} of visual question answering (VQA), is a well-explored problem in the vision literature. 
% Subsequently,VQG

% \newpage
\begin{algorithm}[!t]
\caption{Model Parity Aligner (MPA)}
\label{alg:mpa}
\renewcommand{\algorithmicindent}{0.5em}
\begin{algorithmic}[0]
\State \textbf{Input:} Large Vision Language Model (\lvlm{}) parameterized by $\phi$; Small Vision-Language Model (\svlm{}) parameterized by $\theta$; 
\text{unlabeled images: }$\mathcal{I} = \{I_1, I_2, \cdots, I_N\}$; \textbf{task}: $\mathcal{T}$.
\State \textbf{Output:} Enhanced \svlm{} with updated parameters ($\hat{\theta}$).
\end{algorithmic}
\begin{algorithmic}[1]
    \State $\mathcal{D}_{PA}^{\mathcal{T}}$ $\gets \textbf{PA}(\text{\lvlm{}}_{\phi},  \mathcal{I}, \mathcal{T})$ \Comment{\textcolor{green}{PA - Pseudo Annotator}, $\mathcal{D}_{PA}^{\mathcal{T}}:$ pseudo-annotated data} 
    \State $\mathcal{D}_{PI}^{\mathcal{T}}$ $\gets \textbf{PI}(\text{\lvlm{}}_{\phi}\text{, \svlm{}}_{\theta}\text{, }\mathcal{D}_{PA}^{\mathcal{T}})$ \Comment{\textcolor{blue}{PI - Parity Identifier}, $\mathcal{D}_{PI}^{\mathcal{T}}:$ parity dataset}
    \State \svlm{}$_{\hat{\theta}} \gets \textbf{PL}(\text{\svlm{}$_{\theta}$, $\mathcal{D}_{PI}^{\mathcal{T}}$ })$ \Comment{\textcolor{red}{PL: Parity Leveler}}
    \State \Return \svlm{}$_{\hat{\theta}}$
\end{algorithmic}
\hsize=\columnwidth
\end{algorithm}
\section{\underline{M}odel \underline{P}arity \underline{A}ligner (MPA)}

Given a task $\mathcal{T}$ and a set of unlabeled images $\mathcal{I} =$ $\{{I_i}\}_{i=1}^N$, our goal is to empower small vision language models (\svlm{}s) with task-specific capabilities and improve their performance on the task $\mathcal{T}$. In this work, we restrict ourselves to the VQA task and experiment with various variants of VQA that require interpretation of visual text, chart, and external knowledge. Inspired by the standard machine learning lifecycle~\cite{datasciencecycle}, our proposed \underline{M}odel \underline{P}arity \underline{A}ligner (MPA) framework follows a systematic approach to achieve this goal. The process begins with automatically annotating unlabeled images $\mathcal{I}$ for task $\mathcal{T}$ using the Pseudo Annotator module discussed in Section~\ref{subsec:pa}, followed by strategic data selection with automatic quality assessment of the annotations using the Parity Identification module discussed in Section~\ref{subsec:pi}. This automatically curated and cleaned data is then utilized to fine-tune the \svlm{} model using the Parity Leveler module discussed in Section~\ref{subsec:pl}. The workflow of our proposed MPA framework, which includes its three interconnected modules, is illustrated in Figure~\ref{fig:method} and described in Algorithm~\ref{alg:mpa}. 

The proposed Model Parity Aligner (MPA) consists of three main modules: (a) Pseudo Annotator (PA), (b) Parity Identifier (PI), (c) Parity Leveler (PL). These modules work together to systematically enrich \svlm{}s. The MPA takes \svlm{}$_\theta$, \lvlm{}$_\phi$, a set of unlabeled images ${\mathcal{I}}$ and the task $\mathcal{T}$ as inputs and returns an enhanced \svlm{}$_{\hat{\theta}}$ where $\phi$, ${\theta}$, $\hat{\theta}$ are the parameters of \lvlm{}, \svlm{}, updated \svlm{}, respectively. It should be noted here that $|\hat{\theta}| = |\theta| \ll |\phi|$ where $|\cdot|$ denotes the size of the model. Next, we provide an in-depth overview of each module.

\subsection{Pseudo Annotator (PA)}
\label{subsec:pa}

\begin{algorithm}[!t]
\caption{Pseudo Annotator (PA)}
\label{alg:pa}
\renewcommand{\algorithmicindent}{0.5em}
\begin{algorithmic}[0]
\State \textbf{Input:} Large Vision Language Model (\lvlm{}) parameterized by $\phi$; unlabeled images: $\mathcal{I} = \{I_i\}_{i=1}^N$; task prompt: $\mathcal{T}_{pr}$.
\State \textbf{Output:} pseudo-annotated images for the \textbf{task }$\mathcal{T}$: $\mathcal{D}_{PA}^{\mathcal{T}} = \{(I_i, Q_i, A_i)\}_{i=1}^N$.
\end{algorithmic}
\begin{algorithmic}[1]
    \State $\mathcal{D}_{PA}^{\mathcal{T}} \gets [~~] $
    % \State instruct = Instruction$(\mathcal{T})$
    \For {$I_i \text{ in } \mathcal{I}$}
        \State $(Q, A)_i \gets \text{\lvlm{}}_{\phi}(\mathcal{T}_{pr}, I_i)$
        \State $\mathcal{D}_{PA}^{\mathcal{T}}\text{.append}((I_i,Q_i,A_i))$ \Comment{{Triplet: $(I_i, Q_i, A_i)$ is considered as one pseudo-annotated sample for task $\mathcal{T}$.}}
    \EndFor
    \State \Return $\mathcal{D}_{PA}^{\mathcal{T}}$  
\end{algorithmic}
\hsize=\columnwidth
\end{algorithm}

This module which is described in Algorithm~\ref{alg:pa} is responsible for obtaining pseudo-annotation for unlabeled images $\mathcal{I}$. We employ an \lvlm{} to generate annotations for the unlabeled images for the task $\mathcal{T}$. In this work, we experimented with two \lvlm{}s. Since we have only access to unlabeled images, we ask \lvlm{} to generate task-specific visual question and answer pairs. The generation of visual questions (VQG) has been shown to improve the visio-lingual abilities of a vision and language model~\cite{kafle2017data,kddaug}. In this work, we additionally ask \lvlm{} to generate the corresponding answer. 

To be precise, \lvlm{} is prompted with a task-specific prompt $\mathcal{T}_{pr}$ to create task-specific question-answer pairs\footnote{For example, in the case of ChartQA, $\mathcal{T}_{pr}$ instructs the model to focus on reasoning over charts, including trend analysis and numerical interpretation. Similarly, for TextVQA, the prompt emphasizes reading and comprehending scene text to formulate relevant questions and answers. This ensures that the generated QA pairs align with the specific reasoning challenges posed by each task.} $(Q,A)_i$ for each image $I_i$ within $\mathcal{I}$, where $i \in \{1, \cdots, N\}$. The module produces the pseudo-annotated dataset $\mathcal{D}_{PA}^{\mathcal{T}}$ for task $\mathcal{T}: \{(I_i, Q_i, A_i)\}_{i=1}^N$, with each triplet $(I_i, Q_i, A_i)$ representing an annotated sample for task $\mathcal{T}$. The \lvlm{}-driven automated annotation presents challenges, e.g., (i) noisy annotations and (ii) hallucinated content necessitating careful quality validations. Our proposed PI module, described next, inherently accounts for quality validations and minimizes such noisy annotations, while sampling for parity samples.

\subsection{Parity Identifier (PI)}
\label{subsec:pi}
This module capitalizes on the existing capabilities of \svlm{} while isolating its knowledge gaps relative to \lvlm{}. Rather than following conventional approaches~\cite{kddaug,seltda,changpinyo2022all} of using all pseudo-annotated data for training, we implement a more targeted methodology to identify specific knowledge disparities between models. We evaluated both \lvlm{} and \svlm{} in zero-shot settings by presenting each model with image-question pairs $(I_i, Q_i)$ from the PA-annotated dataset $\mathcal{D}_{PA}^{\mathcal{T}}$. The respective answers - $\tilde{A}_i$ from \lvlm{} and $\hat{A}_i$ from \svlm{}—are then compared against the pseudo annotation $A_i$ using the following expression. 
\begin{equation}
E(X) =
\begin{cases} 
1, & \text{if } X = A, \\
0, & \text{otherwise},
\end{cases}
\quad \text{for } X \in \{\tilde{A}, \hat{A}\}.
\label{eqn:error}
\end{equation}

Further, we select samples that satisfy the following Boolean condition $S$.
\begin{equation}
    % \text{criteria} = E(\tilde{A}) \land \neg E(\hat{A})
    S\big((I, Q, A)\big) =
\begin{cases} 
1, & \text{if } E(\tilde{A}) \land \neg E(\hat{A}), \\
0, & \text{otherwise}.
\end{cases}
\label{eqn:criteria}
\end{equation}
Here, Boolean condition $S$ selects an annotated triplet $(I_i,Q_i,A_i)$ if $\tilde{A}_i$ correctly matches $A_i$ while $\hat{A}_i$ does not, thereby precisely identifying the knowledge gap between the models where \svlm{} requires improvement. In other words, $S$ selects those samples where \lvlm{} answers correctly, while \svlm{} answer is incorrect, assuming the pseudo-annotated answer as ground truth. This methodology inherently performs quality verification by leveraging \lvlm{}'s superior answering capabilities, as these models are primarily instruction-tuned for answering rather than annotating. By selecting only instances where \lvlm{} demonstrates consistency between its annotation and answering phases, PI module effectively filters out noisy or hallucinated annotations. The resulting parity subset $\mathcal{D}_{PI}^{\mathcal{T}}: \{(I_i, Q_i, A_i)\}_{i=1}^K$ with $K \ll N$, constitutes highly efficient samples focused exclusively on the specific knowledge deficiencies of \svlm{}. This targeted approach eliminates the need to train on potentially problematic samples or the entire annotation set, optimizing both training efficiency and model performance. This module is detailed in Algorithm~\ref{alg:pi}.

\begin{algorithm}[!t]
\caption{Parity Identifier (PI)}
\label{alg:pi}
\renewcommand{\algorithmicindent}{0.5em}
\begin{algorithmic}[0]
\State \textbf{Input:} Large Vision Language Model (\lvlm{}) parameterized by $\phi$; Small Vision Language Model (\svlm{}) parameterized by $\theta$; pseudo-annotated data: $\mathcal{D}_{PA}^{\mathcal{T}}$.
\State \textbf{Output:} Parity (Knowledge gap) between \lvlm{} and \svlm{}: $\mathcal{D}_{PI}^{\mathcal{T}} = \{(I_i, Q_i, A_i)\}_{i=1}^K, K \ll N$.
\end{algorithmic}
\begin{algorithmic}[1]
    \State $\mathcal{D}_{PI}^{\mathcal{T}} = [~~]$
    \For{$(I_i, Q_i, A_i)$ in $\mathcal{D}_{PA}^{\mathcal{T}}$}
        \State  $\tilde{A}_i \gets \text{\lvlm{}}_{\phi}(I_i, Q_i)$
        \State $\hat{A}_i \gets \text{\svlm{}}_{\theta}(I_i, Q_i)$
        % \If{$\tilde{A}_i == A_i$ and $\hat{A}_i \neq A_i$} \Comment{Eq.~\ref{eqn:error} \& Eq.~\ref{eqn:criteria}}
        %     \State $\mathcal{D}_{PI}^{\mathcal{T}}\text{.append}((I_i, Q_i, A_i))$ \Comment{Incorrect answer by \svlm{}, while correct answer by \lvlm{}}
        % \Else
        %     \State \textbf{continue}
        % \EndIf
        \If{$\tilde{A}_i == A_i \textbf{ and } \hat{A}_i \neq A_i$} \Comment{Eq.~\ref{eqn:error} \& Eq.~\ref{eqn:criteria}}
    \State $\mathcal{D}_{PI}^{\mathcal{T}}\text{.append}((I_i, Q_i, A_i))$ \Comment{Satisfies Eq.~\ref{eqn:criteria} criteria}
\Else
    \State \textbf{continue}
\EndIf
    \EndFor
    \State \Return $\mathcal{D}_{PI}^{\mathcal{T}}$
\end{algorithmic}
\hsize=\columnwidth
\end{algorithm}

\subsection{Parity Leveler (PL)}
\label{subsec:pl}
This module fine-tunes \svlm{} on the parity (knowledge gap) samples identified by the PI module for $L$ number of iterations. We feed each sample $\{I, Q\}_i$ from $\mathcal{D}_{PI}^{\mathcal{T}}$, within an instruction prompt template to \svlm{} to generate the accurate answer $A_i$ to the visual question $Q_i$ on the image $I_i$. \svlm{} learns $P(A_i|Q_i, I_i)$ by modeling the task as a text generation problem, auto-regressively generating the tokens in the answer.
\begin{equation}
\small
    \label{eqn:loss_vqa}
    \mathcal{L}_{gen}(\theta) =  -\frac{1}{b} \sum_{i=1}^{b}\left[\sum_{t=1}^{m}\log P_\theta(A_{i_t}|A_{i<t}, \{I_i, Q_i\})\right]
\end{equation}

Once all answer tokens $A_{i_{1:m}}$ are obtained, we optimize the model using the generation loss $\mathcal{L}_{gen}$, defined over the minibatches of size $b$ samples (Eq.\ref{eqn:loss_vqa}) which is minimized via stochastic gradient descent. Note that \lvlm{} parameters $\phi$ remain frozen throughout MPA.
For an algorithmic description of this module, refer to Algorithm~\ref{alg:pl}.

\begin{algorithm}[!t]
\caption{Parity Leveler (PL)}
\label{alg:pl}
\renewcommand{\algorithmicindent}{0.5em}
\begin{algorithmic}[0]
\State \textbf{Input:} Small Vision-Language Model (\svlm{}) parameterized by $\theta$; parity set: $\mathcal{D}_{PI}^{\mathcal{T}} = \{(I_i, Q_i, A_i)\}_{i=1}^K$
\State \textbf{Output:} Enhanced \svlm{} with updated parameters $\hat{\theta}$
\end{algorithmic}

\begin{algorithmic}[1]
\For{iter $= 1$ to $L$} \Comment{$L$: total no. of iterations}
    \For{$\{(I_i, Q_i, A_i)\}_{i=1}^b$ in $\mathcal{D}_{PI}^{\mathcal{T}}$} \Comment{$b$: batch size}
        \State $\{\hat{A}_i\}_{i=1}^b \gets \text{\svlm{}}_\theta(\{(I_i, Q_i)\}_{i=1}^b)$
        \State Compute $\mathcal{L}_{gen}(\{\hat{A}_i, A_i\}_{i=1}^b)$ \Comment{Answer generation loss}
        \State Update $\theta$ using $\mathcal{L}_{gen}$ \Comment{Gradient descent}
    \EndFor
\EndFor
\State \Return \svlm{}$_{\hat{\theta}}$
\end{algorithmic}
\hsize=\columnwidth
\end{algorithm}

\section{Experiments and Results}

\begin{table*}[]
\centering
\resizebox{\textwidth}{!}
{
\begin{tabular}{l c cccc cccc c c}
\hline
 & & \multicolumn{8}{c}{\lvlm{}} & \multicolumn{2}{c}{Gains} \\
 \cmidrule(r){3-10} \cmidrule(r){11-12}
 & & \multicolumn{4}{c}{Qwen2VL-7B~\cite{qwen2vl}} & \multicolumn{4}{c}{InternVL2-8B~\cite{chen2024internvl}} & \multirow{2}{*}{Max} & \multirow{2}{*}{Average} \\
 \cmidrule(r){3-6}
 \cmidrule(r){7-10}
{\svlm{}} & Method  & TextVQA & ST-VQA & ChartQA & OKVQA & TextVQA & ST-VQA & ChartQA & OKVQA & & \\
\hline
\multirow{2}{*}{SmolVLM-500M} & ZS & 55.3 & 78.5 & 56.5 & 38.2 & 55.3 & 78.5 & 56.5 & 38.2 & \multirow{2}{*}{3.4} & \multirow{2}{*}{2.4} \\
 & \textbf{MPA} & \textbf{57.6}$_{(+2.3)}$ & \textbf{80.3}$_{(+1.8)}$ & \textbf{59.9}$_{(+3.4)}$ & \textbf{40.7}$_{(+2.5)}$ & \textbf{57.7}$_{(+2.4)}$ & \textbf{80.7}$_{(+2.2)}$ & \textbf{59.3}$_{(+2.8)}$ & \textbf{39.9}$_{(+1.7)}$ & & \\
\hline
\multirow{2}{*}{TinyLLaVA-2B} & ZS & 47.1 & 44.7 & 12.0 & 43.6 & 47.1 & 44.7 & 12.0 & 43.6 & \multirow{2}{*}{15.2} & \multirow{2}{*}{6.8} \\
 & \textbf{MPA} 
 & \textbf{53.5}$_{(+6.4)}$ & \textbf{48.7}$_{(+4.0)}$
 & \textbf{24.0}$_{(+12.0)}$ & \textbf{46.6}$_{(+3.0)}$ & \textbf{51.9}$_{(+4.8)}$ & \textbf{49.8}$_{(+5.1)}$ & \textbf{27.2}$_{(+15.2)}$ & \textbf{47.2}$_{(+3.6)}$ & & \\
 \hline
\multirow{2}{*}{InternVL2-2B} & ZS & 68.0 & 63.0 & 63.2 & 42.7 & 68.0 & 63.0 & 63.2 & 42.7 & \multirow{2}{*}{5.1} & \multirow{2}{*}{3.0} \\
 & \textbf{MPA} & \textbf{70.3}$_{(+2.3)}$ & \textbf{65.5}$_{(+2.5)}$ & \textbf{68.3}$_{(+5.1)}$ & \textbf{45.6}$_{(+2.9)}$ & \textbf{69.5}$_{(+1.5)}$ & \textbf{65.7}$_{(+2.7)}$ & \textbf{68.2}$_{(+5.0)}$ & \textbf{44.6}$_{(+1.9)}$ & & \\
\hline
\multirow{2}{*}{InternVL2-4B} & ZS & 69.1 & 63.2 & 73.1 & 50.5 & 69.1 & 63.2 & 73.1 & 50.5 & \multirow{2}{*}{4.7} & \multirow{2}{*}{2.1} \\
 & \textbf{MPA} & \textbf{71.4}$_{(+2.3)}$ & \textbf{66.6}$_{(+3.4)}$ & \textbf{73.8}$_{(+0.7)}$ & \textbf{52.3}$_{(+1.8)}$ & \textbf{70.3}$_{(+1.2)}$ & \textbf{67.9}$_{(+4.7)}$ & \textbf{74.0}$_{(+0.9)}$ & \textbf{52.0}$_{(+1.5)}$ & & \\
\hline
\multirow{2}{*}{Qwen2VL-2B} & ZS & 70.6 & 62.5 & 65.9 & 47.1 & 70.7 & 62.5 & 65.9 & 47.1 & \multirow{2}{*}{4.7} & \multirow{2}{*}{2.6} \\
 & \textbf{MPA} & \textbf{75.1}$_{(+4.5)}$ & \textbf{67.2}$_{(+4.7)}$ & \textbf{67.6}$_{(+1.7)}$ & \textbf{48.9}$_{(+1.8)}$ & \textbf{72.3}$_{(+1.6)}$ & \textbf{66.6}$_{(+4.1)}$ & \textbf{66.9}$_{(+1.0)}$ & \textbf{48.9}$_{(+1.8)}$ & & \\
\hline
\end{tabular}}
\caption{Comparison of our proposed MPA framework performance with the baselines on TextVQA, ST-VQA, ChartQA and OKVQA. The parenthesis (+x) denotes the improvement of +x\% over the zero-shot \svlm{} by our proposed MPA. The max and average columns show the overall performance gains across all tests for each \svlm{}.}
\label{tab:model-differences}
\end{table*}

\begin{figure*}[!t]
\centering
  \includegraphics[width=\textwidth]{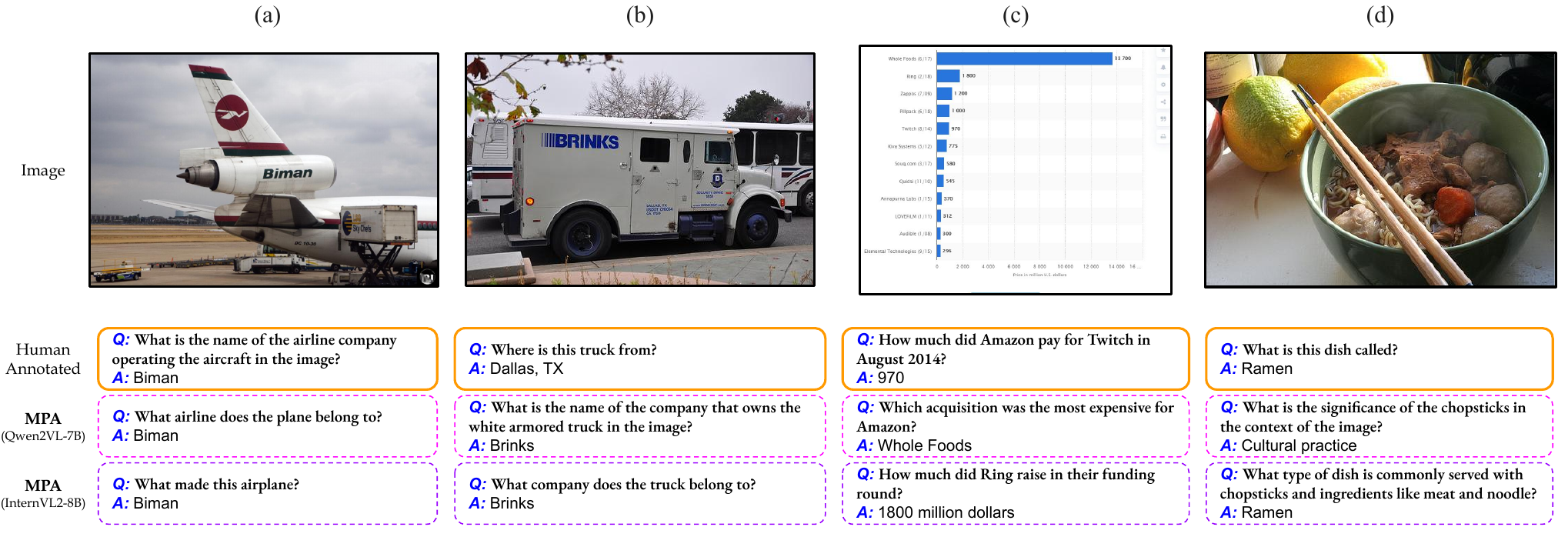}
  \caption{A selection of few pseudo annotations generated by our framework. We further show human annotations from their respective original dataset train splits. \textbf{(Best viewed in color)}.}
  \label{fig:visRes1}
\end{figure*}

% \subsection{Datasets}
\noindent \textbf{Datasets.} We evaluate our approach on four widely-used public VQA benchmarks, namely, TextVQA~\cite{textvqa}, ST-VQA~\cite{stvqa}, ChartQA~\cite{masry2022chartqa}, and OKVQA~\cite{okvqa}. These datasets are relevant to MPA because they introduce diverse reasoning challenges, such as text, chart, external world understanding beyond traditional VQA~\cite{antol2015vqa}, making them strong benchmarks for evaluating gains in \svlm{}. More details on these datasets are in Appendix~\ref{sec:app_dataset}. Further, as MPA is primarily designed for label-free training, we exclude all question-answer annotations from the training splits of each dataset during evaluation.

\noindent \textbf{S-VLMs and L-VLMs used.}
Following the parameter-based taxonomy defined for Vision-Language Models (VLMs) in Section~\ref{sec:related_work}, where models with $\leq$ 5B parameters are classified as small VLMs (\svlm{}s), while those exceeding 5B parameters are large VLMs (\lvlm{}s)~\cite{lu2024small}, we chose five models that range from 500M to 4B parameters as \svlm{}, namely SmolVLM-500M~\cite{smolvlm}, TinyLLaVA-2B~\cite{zhou2024tinyllava}, InternVL2-2B~\cite{chen2024internvl}, Qwen2VL-2B~\cite{qwen2vl}, and InternVL2-4B~\cite{chen2024internvl}; and two open-source models viz. Qwen2VL-7B~\cite{qwen2vl} and InternVL2-8B~\cite{chen2024internvl} and one closed-source model, i.e., GPT-4o~\cite{openai2024gpt4o} as \lvlm{}.

\subsection{Results and Discussion}
\label{sec:rnd}
We present the quantitative results of our MPA framework across four datasets evaluated in ten combinations of two \lvlm{}s and five \svlm{}s in Table~\ref{tab:model-differences}. The results show that MPA consistently improves the performance of all \svlm{}s in all datasets with 15.2\% maximum and 3.4\% average gain in an absolute scale. Here, we analyze the results from the following three key perspectives.

\noindent\textbf{(i) \svlm{} family-specific analysis} The most noticeable gains are as follows (refer Table~\ref{tab:model-differences}): TinyLLaVA-2B achieves 27.2\% accuracy on ChartQA with our MPA framework, guided by InternVL2-8B, marking an absolute improvement of +15.2\% over its original zero-shot performance. Similarly, Qwen2VL-2B, guided by Qwen2VL-7B and InternVL2-4B, guided by InternVL2-8B in our MPA framework achieve +4.7\% and +4.7\% improvements, respectively, on ST-VQA. On ChartVQA, SmolVLM-500M, guided by Qwen2VL-7B in our MPA framework, improves by +3.4\%, while InternVL2-2B, guided by Qwen2VL-7B, gains +5.1\%. These results highlight effectiveness of MPA in enhancing the performance of \svlm{}s across diverse VQA tasks.

\begin{table}[t!]
\centering
\resizebox{0.7\columnwidth}{!}
{
\begin{tabular}{l c}
\toprule
\svlm{} & GPT-4o as \lvlm{} \\

\hline

TinyLLaVA-2B & 47.1 \\
\rowcolor{gray!10} TinyLLaVA-2B + MPA & \textbf{55.4}$_{(+8.3)}$ \\
\hline
Qwen2VL-2B & 70.6 \\
\rowcolor{gray!10} Qwen2VL-2B + MPA & \textbf{75.4}$_{(+4.8)}$ \\
\hline
\end{tabular}
}
\caption{Comparison of MPA-aligned \svlm{}s against baseline \svlm{}s on TextVQA, with GPT-4o as LVLM. 
}
\label{tab:close_source}
\end{table}

\noindent\textbf{(ii) VQA Task-specific analysis} 
We observe that TinyLLaVA-2B+MPA aligned with InternVL2-8B achieves a notable +15.2\% gain on ChartQA, highlighting our MPA's strength as a knowledge alignment module. In this scenario, it effectively identifies and bridges the knowledge gap between \lvlm{} and \svlm{} for `\textit{complex visual reasoning that involves interpreting charts and graphs}. Improvements on TextVQA (+6.4\%) and ST-VQA (+5.1\%) further demonstrate MPA’s ability to transfer `\textit{visual text understanding}' from larger to smaller models. The modest gain on OKVQA reflects its reliance on external knowledge, which \svlm{} inherently lack. While MPA enhances internal knowledge utilization, it cannot fully address such gaps without RAG or fine-tuning on knowledge-rich data. The results validate the effectiveness of MPA within its scope, while highlighting the challenges of knowledge-intensive visual question answering.

\begin{table}[t!]
\centering
\resizebox{\columnwidth}{!}
{
\begin{tabular}{lcccc}
\hline
\textbf{Task} & \textbf{Dataset} & \textbf{Metric} & \textbf{SVLM} & \textbf{SVLM+MPA} \\
\hline
OCR & ICDAR2015~\cite{ICDAR2015} & WRR & 31.9 & \textbf{36.4}($\uparrow 4.5$) \\
\hline
\multirow{3}{*}{TC} & \multirow{3}{*}{TextCaps~\cite{sidorov2020textcaps}} & BLEU-1 & 7.9 & \textbf{15.3}($\uparrow 7.4$) \\
 &  & ROUGE-L & 17.4 & \textbf{20.6}($\uparrow 3.2$) \\
 &  & CIDEr & 8.7 & \textbf{38.1}($\uparrow 29.4$) \\
\hline
\end{tabular}}
\caption{\textbf{MPA transfers the fundamental capabilities beyond VQA.} In our MPA framework, we use \svlm{}: TinyLLaVA-2B, \lvlm{}: Qwen2VL-7B. Here, \textbf{OCR:} visual-text recognition, \textbf{TC:} text-aware image captioning. \textbf{WRR:} word recognition rate.}
\label{tab:text_understanding}
\end{table}

\noindent\textbf{(iii) Model size-specific Analysis:} MPA improves performance on all model scales, from SmolVLM-500M to InternVL2-4B, demonstrating its versatility. In particular, TinyLLaVA-2B achieves the highest average gain of +6.8 across all tasks, whereas InternVL2-4B shows a comparatively modest improvement of +2.1. We attribute this contrast to two factors: (i) Pretraining data gaps: smaller models like TinyLLaVA-2B benefit more from MPA as it effectively fills missing capabilities through targeted alignment; (ii) Diminishing returns with scale: it is inherently harder to align larger models (4B in this case) that already possess stronger capabilities, in line with scaling laws.

\noindent \textbf{(iv) \lvlm{}-Specific Analysis:}
We analyze the effectiveness of different guiding \lvlm{}s within MPA by computing average gains across five \svlm{}s and four VQA datasets. Qwen2VL-7B achieves the highest average improvement of +3.5 points, followed closely by InternVL2-8B with +3.2 points. This suggests that while both models are effective guides, Qwen2VL-7B offers a slightly stronger alignment signal, potentially due to differences in their pretraining objectives or representations. These results highlight that MPA is robust to the choice of \lvlm{}, yet benefits from stronger or more task-aligned guides.

\subsubsection{Ablations and Analysis}
\label{sec:aa}
We conduct the following ablations and analysis:% of the proposed work:

\noindent\textbf{\textit{(i) How effective is MPA in aligning \svlm{}s with closed-source models?:}}
MPA can also leverage powerful closed-source \lvlm{}s to improve \svlm{}s. 
To assess this, we performed experiments using GPT-4o~\cite{openai2024gpt4o} as the guiding \lvlm{}. As shown in Table~\ref{tab:close_source}, MPA consistently improves performance across all aligned \svlm{}s, despite having no access to the guiding model's logits or weights. This demonstrates MPA’s unique advantage over standard distillation methods, which require full model access. With the expected rise in powerful closed-source models~\cite{openai2024gpt4o,team2023gemini}, such alignment strategies become increasingly valuable. In fact, our results show that integrating powerful \lvlm{}, e.g. GPT-4o through MPA brings \svlm{}s closer or even better in performance to significantly larger models, e.g., MPA-aligned Qwen2VL-2B (75.4\%) outperforms Qwen2VL-7B (74.7\%). 

\noindent\textbf{\textit{(ii) Does MPA transfers the fundamental capabilities beyond VQA?:}} 
MPA is designed to enhance the VQA performance of \svlm{}s by aligning them with \lvlm{}s, and our results confirm its effectiveness. To examine whether MPA also transfers broader fundamental capabilities such as visual text understanding, we evaluate zero-shot TinyLLaVA-2B and its MPA-aligned counterpart on two different tasks: visual text recognition on ICDAR 2015~\cite{ICDAR2015} and text-aware image captioning on  TextCaps~\cite{sidorov2020textcaps}, using Qwen2VL-7B as the guiding \lvlm{} in MPA. As shown in Table~\ref{tab:text_understanding}, the MPA-aligned model improved text recognition accuracy by 4.5\% on an absolute scale and yields notable improvements in captioning metrics such as ROUGE-L and CIDEr. These results suggest that MPA transfers fundamental text understanding capabilities from \lvlm{}s to \svlm{}s beyond the VQA.

\begin{table}[t!]
    \centering
    \resizebox{0.9\columnwidth}{!}
    {
    \begin{tabular}{l  l  c c c c }
    \hline
         {\lvlm{}} & Status & A $\uparrow$ & AC $\uparrow$ & TR $\uparrow$ & HLS $\uparrow$\\
         \hline
         \multirow{2}{*}{Qwen2VL-7B} & Pre-PI & 0.76 & 0.68 & 0.8 & 58 \\
         & Post-PI & \textbf{0.92} & \textbf{0.84} & \textbf{0.92} & \textbf{74} \\
         \hline
         \multirow{2}{*}{InternVL2-8B} & Pre-PI & 0.74 & 0.65 & 0.78 & 56 \\
         & Post-PI & \textbf{0.87} & \textbf{0.78} & \textbf{0.88} & \textbf{73} \\
         \hline
    \end{tabular}
    }
    \caption{User study on the pseudo-annotations quality: Pre-PI and Post-PI in MPA. \textbf{A}: answerability, \textbf{AC}: answer correctness, \textbf{TR}: task relevancy, \textbf{HLS}: Human Likeness Score. Refer Section~\ref{sec:aa} for more details.}
    \label{tab:textvqg_main_results}
\end{table}

\noindent\textbf{\textit{(iii) How effective is the role of PI in pseudo-annotation quality correction?:}} Incorrect annotations may cause models to learn spurious patterns, exhibit biased behavior, and suffer from degraded performance and reliability in downstream tasks. To assess the impact of the PI module on genereted annotation quality, we conducted a user study in which three annotators evaluated 500 randomly sampled pseudo-annotations prior and post PI processing. The evaluation used the following metrics: (a) \emph{Answerability (A)}: 1 if the question is answerable from the image, 0 otherwise; (b) \emph{Answer Correctness (AC)}: 1 if the answer is correct, assuming the question is valid; (c) \emph{Task Relevance (TR)}: 1 if the question aligns with the task, 0 otherwise; and (d) \emph{Human-Likeness Score (HLS)}: percentage of PI-sampled annotations mistaken for human-annotated ones in a mixed set. As shown in Table~\ref{tab:textvqg_main_results}, post-PI annotations exhibited higher quality across all metrics, with more being identified as human-annotated. Figure~\ref{fig:visRes1} provides visual evidence by illustrating the high correlation between MPA-generated annotations and human annotated samples. These results validate that PI effectively filters noise and corrects errors, enhancing the overall reliability of MPA-generated annotations.

\begin{table}[t!]
\centering
\resizebox{0.8\columnwidth}{!}
{
\begin{tabular}{l cccc}
\hline
Method  & TextVQA & ST-VQA & ChartQA & OKVQA \\
\hline
LoRA SFT & 71.9 & 63.4 & 66.1 & 47.9 \\
Full SFT & 71.8 & 61.7 & 65.7 & 47.7 \\
MPA & \textbf{75.1} & \textbf{67.2} & \textbf{67.6} & \textbf{48.9} \\
\hline
\end{tabular}}
\caption{Comparison of few-shot methods vs MPA-aligned Qwen2VL-2B with Qwen2VL-7B as \lvlm{}. Please note that MPA operates without any human-labeled samples, whereas the other two baselines each use 100 human-labeled samples.}
\label{tab:sft-lora-results}
\end{table}

\noindent\textbf{\textit{(iv) How does MPA compare to few-shot supervised baselines?}} While MPA is designed for a setting where human-labeled traning data is unavailable, obtaining a small labeled set (e.g., 100 samples) is often feasible. In such scenarios, commonly adopted few-shot supervised methods like LoRA-based SFT and full SFT can be applied directly to the \svlm{}. To benchmark MPA against these methods, we fine-tune Qwen2VL-2B using both approaches and compare them with MPA-aligned Qwen2VL-2B (using Qwen2VL-7B as \lvlm{}). As shown in Table~\ref{tab:sft-lora-results}, MPA consistently outperforms both baselines without labeled supervision, demonstrating high-quality label generation and effective knowledge transfer.

\noindent\textbf{\textit{(v) Does PI filtering improve over raw pseudo-labels or full human-labeled data?}}
While our primary focus is on label-free training using MPA, we further investigate the quality of supervision introduced by PI filtering. Specifically, we compare three settings for training Qwen2VL-2B: (i) full human-labeled data (HL), (ii) pseudo-labeled data (PL) from  MPA without PI filtering, and (iii) high-quality subset selected by PI that targets the knowledge gap. As shown in Table~\ref{tab:pi_ablation}, the PI-selected subset achieves the highest accuracy across all tasks--TextVQA (75.1\%), ST-VQA (67.2\%), and ChartQA (67.6\%), despite using far fewer samples. Interestingly, the performance gain from full human-labeled data over zero-shot baselines is relatively limited. Prior work~\cite{seltda} suggests that excessive labeled data can introduce redundancy or noise, reducing the marginal benefit of supervision. This highlights the value of PI filtering in identifying high-utility samples that yield more efficient and effective learning. 

\begin{table}[t!]
  \centering
  \resizebox{\columnwidth}{!}
  {
  \begin{tabular}{l c c c c c c c}
  \toprule
    \multirow{2}{*}{Data} & \multirow{2}{*}{Labels} & \multicolumn{2}{c}{TextVQA} & \multicolumn{2}{c}{ST-VQA} & \multicolumn{2}{c}{ChartQA} \\
    \cmidrule(lr){3-4} \cmidrule(lr){5-6} \cmidrule(lr){7-8}
    & & \#Samples $\downarrow$ & Acc. $\uparrow$ & \#Samples $\downarrow$ & Acc. $\uparrow$ & \#Samples $\downarrow$ & Acc. $\uparrow$ \\
    \midrule
     Original & HL & 35K & 72.7 & 22K & 65.5 & 28K & 66.9 \\
     % MPA (w/o PI)$^*$ & PL & 21K & 73.3 & 15K & 65.5 & 19K & 67.4 \\
     MPA (w/o PI) & PL & 21K & 73.6 & 15K & 65.8 & 19K & 67.4 \\
     MPA & PL & \textbf{2K} & \textbf{75.1} & 1.5K & \textbf{67.2} & 1.6K & \textbf{67.6} \\
    \bottomrule
    \end{tabular}
    }
    \caption{Ablation result of using samples from MPA v/s MPA without PI filtering, with Qwen2VL-7B as \lvlm{} and Qwen2VL-2B as \svlm{} inside MPA. HL: Human Labeled. PL: Pseudo Labeled.}
    \label{tab:pi_ablation}
\end{table}
% $^*$: Answers generated by the L-VLM in PI phase ($\tilde{A}$) are used instead of the answers generated in the PA phase ($A$).

\begin{table}[t]
\centering
\resizebox{0.7\columnwidth}{!}
{
\begin{tabular}{llc}
\toprule
\textbf{Model} & Method & \textbf{Acc. (\%)} \\
\midrule
TinyLLaVA-2B & ZS & 51.2 \\
TinyLLaVA-2B & MPA & \textbf{53.6}$_{(+2.4)}$ \\
\bottomrule
\end{tabular}}
\caption{Performance on Medical VQA (PathVQA). MPA-aligned TinyLLaVA-2B (with Qwen2VL-7B as \lvlm{}) shows improved cross-domain generalization.}
\label{tab:medical_vqa}
\end{table}

\noindent\textbf{\textit{(vi) Beyond standard VQA applicability (Medical VQA):}}
To evaluate MPA’s utility beyond standard VQA tasks, we assess its performance in the medical domain using the PathVQA dataset~\cite{he2020pathvqa}. We compare zero-shot TinyLLaVA-2B with its MPA-aligned counterpart, guided by Qwen2VL-7B. We focus on the binary (yes/no) subset of PathVQA, as the open-ended questions often contain highly specialized medical terminology that poses challenges even for large models and may not reflect generalizable reasoning capabilities. As shown in Table~\ref{tab:medical_vqa}, MPA yields a gain of +2.4\%, demonstrating effective knowledge transfer even in diverse domain-specific settings. These results highlight MPA’s ability to generalize across domains without requiring task-specific data or fine-tuning.

\noindent \textit{\textbf{(vii) Knowledge Gap Analysis.}} To better characterize the nature of the ``knowledge gaps'' between \textsc{S-VLM}s and \textsc{L-VLM}s, we manually inspected 100 randomly selected $\mathcal{D}_{PI}$ samples per task. In particular, we compared Qwen2VL-2B and Qwen2VL-7B within the MPA framework. We categorize the dataset-wise knowledge gaps into key categories, which are summarized in Table~\ref{tab:knowledgeGaps}. Note that \textit{Correct but Verbose} and \textit{Noisy/Task-Irrelevant} cases are excluded from the knowledge-gap categories, as they do not represent fundamental reasoning shortcomings.

Further, we provide representative examples that illustrate these knowledge-gap categories:  

\begin{enumerate}
    \item \textbf{TextVQA/ST-VQA: }  
    % \begin{itemize}
        (i) \textit{Shallow OCR grounding} (Fig.~\ref{fig:visRes2} (a)): ``What word is printed under interior design on the book in the middle?'' --- \textsc{S-VLM} outputs ``para'' from a nearby visible region instead of grounding to the queried location.  
        (ii) \textit{Noisy or hallucinated OCR} (Fig.~\ref{fig:stvqa_addn_results} (d)): ``What company’s logo is in the black box in the upper left?'' --- \textsc{S-VLM} hallucinates ``Burberry'' without actually reading the text.  
    % \end{itemize}

    \item \textbf{ChartQA: }  
    % \begin{itemize}
        (i) \textit{Entity misalignment} (Fig.~\ref{fig:chartvqa_addn_results} (d)): ``Who was the leading goal scorer for Celtic FC as of September 2020?'' --- the \textsc{S-VLM} retrieves an incorrect player name that is not aligned with the queried entity.  
        (ii) \textit{Conditional/chart understanding error} (Fig.~\ref{fig:visRes2} (c)): ``Which year yielded the smallest difference between men and women students?'' --- \textsc{S-VLM} fails to detect the year with the minimum gap between the trend lines.  
        (iii) \textit{Trend misinterpretation} (Fig.~\ref{fig:chartvqa_addn_results} (b)): ``Does the life expectancy decrease over the years?'' --- \textsc{S-VLM} misinterprets the slope changes.  
    % \end{itemize}

    \item \textbf{OKVQA}  
    % \begin{itemize}
        (i) \textit{Lack of internal knowledge grounding} (Fig.~\ref{fig:aokvqa_addn_results} (b)): ``At what speed does this animal run?'' --- \textsc{S-VLM} fails to answer, while the MPA-aligned model succeeds without external knowledge, highlighting the shallow grounding of the \textsc{S-VLM}.  
        (ii) \textit{Visual guesswork} (Fig.~\ref{fig:aokvqa_addn_results} (c)): ``What is the name of the floor pattern?'' --- \textsc{S-VLM} guesses ``diamond'' from vague cues instead of leveraging the consistent checkered pattern.  
    % \end{itemize}
\end{enumerate}

\begin{table}[t]
\centering
\small

\resizebox{\columnwidth}{!}{
\begin{tabular}{l l c}
\hline
\textbf{Dataset} & \textbf{Error Category} & \textbf{\# Samples} \\
\hline
\multirow{4}{*}{TextVQA/ST-VQA} 
& Shallow OCR grounding & 33 \\
& Noisy or hallucinated OCR & 53 \\
& Correct but Verbose & 5 \\
& Noisy/Task-Irrelevant samples & 9 \\
\hline
\multirow{5}{*}{ChartQA} 
& Entity misalignment & 55 \\
& Conditional/chart understanding errors & 17 \\
& Trend misinterpretation & 14 \\
& Correct but Verbose & 8 \\
& Noisy/Task-Irrelevant samples & 6 \\
\hline
\multirow{4}{*}{OKVQA} 
& Lack of internal knowledge grounding & 23 \\
& Visual guesswork & 58 \\
& Correct but Verbose & 4 \\
& Noisy/Task-Irrelevant samples & 15 \\
\hline
\end{tabular}
}
\caption{Distribution of error categories across datasets in our manual inspection of 100 $\mathcal{D}_{PI}$ samples per task. Note that \textit{Correct but Verbose} and \textit{Noisy/Task-Irrelevant} are not true knowledge-gap categories.}
\label{tab:knowledgeGaps}
\end{table}

\section{Conclusion and Future Work}
In this work, we introduced the Model Parity Aligner (MPA), a novel framework that enhances small vision-language models (\svlm{}s) by leveraging unlabeled images and effective knowledge transfer from large vision-language models (\lvlm{}s). Unlike traditional knowledge distillation techniques that rely on labeled data and access to large model logits, MPA employs pseudo-labeling with quality assessment, ensuring that small models learn from high-confidence supervision while avoiding error propagation. Our experiments across four diverse VQA benchmarks, viz. TextVQA, ST-VQA, ChartQA and OKVQA demonstrate that MPA consistently improves \svlm{} performance, making them more viable for real-world applications with limited resources.

Despite these improvements, there still remains a gap between \svlm{}s and \lvlm{}s that highlights the need for further advancements. As future work, we aim to explore more robust knowledge alignment strategies, including iterative refinement of pseudo-labels, leveraging diverse sources of unlabeled data, and integrating multi-step reasoning from \lvlm{}s into \svlm{}s training. Additionally, extending MPA to tasks beyond visual question answering could further enhance its applicability. We view MPA as a first step toward achieving model parity in vision and language models via targeted knowledge alignment, and firmly believe that it shall open up future research avenues for more efficient and capable small models for vision-language tasks.

\section*{Limitations}
Our proposed MPA framework depends on access to a large vision-language model (\lvlm{}) for generating and validating pseudo-annotations. In even stricter resource-constrained settings, this may limit applicability of MPA. Further, when leveraging proprietary closed-source models via commercial APIs, reproducibility and transparency may be compromised due to limited insight into model behavior and potential changes in API responses over time. Our experiments also focus primarily on English-language datasets and VQA-related tasks; generalization to multilingual, or more complex reasoning tasks remains an open direction.

\section*{Ethical Considerations and Broader Impact}
In this work, we used open-source datasets which may contain social or cultural biases. The proposed framework also depends on outputs from large-scale vision-language models (\lvlm{}s), which are known to occasionally generate hallucinated or biased content. Although the Parity Identifier (PI) module is designed to filter out low-quality or incorrect annotations, it cannot entirely eliminate inherited biases from the underlying \lvlm{}. Further, this work involves a human evaluation study in which three annotators were employed to assess the quality of pseudo-annotations generated by our MPA framework. All annotators were compensated fairly in accordance with local wage norms. They were not exposed to harmful, offensive, or sensitive content, and no personally identifiable information was collected at any stage of the study.

\noindent \textbf{Broader Impact}: The proposed MPA framework enables efficient training of small vision-language models (\svlm{}s) using only unlabeled data, reducing reliance on expensive human annotations. By transferring capabilities from large vision-language models (\lvlm{}s) to compact models, MPA makes high-performing multimodal systems more accessible in low-resource settings. This democratization of vision-language technology can benefit real-world applications in healthcare, agriculture, and accessibility, particularly in regions with limited compute or labeled data. Furthermore, the proposed approach encourages the development of scalable alignment strategies that can generalize to diverse, resource-constrained communities.

% Since December 2023, a "Limitations" section has been required for all papers submitted to ACL Rolling Review (ARR). This section should be placed at the end of the paper, before the references. The "Limitations" section (along with, optionally, a section for ethical considerations) may be up to one page and will not count toward the final page limit. Note that these files may be used by venues that do not rely on ARR so it is recommended to verify the requirement of a "Limitations" section and other criteria with the venue in question.

\section*{Acknowledgments}
This work was partly supported by the National Language Translation Mission (NLTM): Bhashini project by the MeitY, Government of India. Abhirama Subramanyam Penamakuri was supported by the PMRF fellowship, MoE, Government of India.

\bibliography{custom}
\newpage
\clearpage

\appendix

% \section{Appendix}

% This is an appendix.
\section{Additional Analysis}

\noindent \textbf{\textit{(i) Additional comparisons: utility of PI filtering over raw pseudo-labels.}}
We extend the analysis of PI filtering by reporting the results of MPA (w/o PI) across all \svlm{}s with Qwen2VL-7B as the guiding \lvlm{} within MPA. As shown in  Table~\ref{tab:pi_ablation_additional}, MPA consistently outperforms MPA (w/o PI) across all tasks and \svlm{}s, despite using far fewer training samples (for instance, $\sim$2K vs.\ $\sim$21K for TextVQA). These results reinforce the utility of PI filtering in isolating knowledge-gap samples that provide more efficient and targeted supervision.

\noindent \textbf{\textit{(ii) Expanded comparison on OCR and text-aware captioning tasks.}}
In Table~\ref{tab:text_understanding}, we examined whether MPA-trained models can improve fundamental capabilities such as OCR and text-aware image captioning, even without direct supervision. We further evaluate this setting by comparing against models fine-tuned on the original human-labeled training splits of TextVQA; the results are presented in Table~\ref{tab:ocr_tc_appendix}. As shown, MPA not only improves over the zero-shot baseline but also surpasses models trained with human-labeled annotations. This highlights that the gains stem from the effectiveness of the MPA pipeline, rather than from overlap between benchmarks, and demonstrates that MPA successfully transfers core visual-linguistic capabilities in a label-free manner.

\noindent \textit{\textbf{(iv) Computational and API cost of PA and PI: }}
MPA is a one-time pipeline where each image is processed by the \lvlm{} during the PA phase, and each generated (image, question) pair is passed once through the \lvlm{} and \svlm{} during the PI phase. For open-source \lvlm{}s like Qwen2VL-7B deployed locally, this is computationally lightweight: on a machine with 3 A6000 (48GB) GPUs, generating approximately 21K pseudo-annotations (e.g., for TextVQA) takes around 4-6 hours end-to-end. Further, the PI step takes another 2-3 hours to identify the samples that represent the knowledge gaps. Alternatively, while using GPT-4o via API, we estimate the total cost of PA + PI for a single \svlm{}–task pair to be around \$11, making MPA a highly cost-effective label-free alternative to supervised training.

\begin{table}[t!]
  \centering
  \resizebox{1\columnwidth}{!}
  {
  \begin{tabular}{l c c c c c}
  \toprule
    \svlm{} & Samples & TextVQA & ST-VQA & ChartQA & OKVQA\\
    \midrule
    \multirow{2}{*}{TinyLLaVA-2B} & MPA (w/o PI) & 56.4 & 79.1 & 57.5 & 39.2 \\
    & MPA & \textbf{57.6} & \textbf{80.3} & \textbf{59.9} & \textbf{40.7} \\
    \midrule
    \multirow{2}{*}{TinyLLaVA-2B} & MPA (w/o PI) & 52.1 & 46.3 & {23.3} & 44.6 \\
    & MPA & \textbf{53.5} & \textbf{48.7} & \textbf{24.0} & \textbf{46.6} \\
    \midrule
    \multirow{2}{*}{InternVL2-2B} & MPA (w/o PI) & 69.0 & 64.5 & 66.7 & 44.0 \\
    & MPA & \textbf{70.3} & \textbf{65.5} & \textbf{68.3} & \textbf{45.6} \\
    \midrule
    \multirow{2}{*}{InternVL2-4B} & MPA (w/o PI) & 69.8 & 65.1 & 72.9 & 51.2 \\
    & MPA & \textbf{71.4} & \textbf{66.1} & \textbf{73.8} & \textbf{52.3} \\
    \midrule
    \multirow{2}{*}{Qwen2VL-2B} & MPA (w/o PI) & 73.6 & 65.8 & 67.4 & 47.2 \\
    & MPA & \textbf{75.1} & \textbf{67.2} & \textbf{67.6} & \textbf{48.9} \\
    \bottomrule
    \end{tabular}
    }
    \caption{Additional results for MPA vs. MPA (w/o PI) across all S-VLMs, using Qwen2VL-7B as the \lvlm{} inside MPA.}
    \label{tab:pi_ablation_additional}
\end{table}

\section{Dataset Details}
\label{sec:app_dataset}
TextVQA consists of 28K images with 45K manually annotated question-answer pairs. It is split into 21K images with 35K questions for training, 3K images with 3.7K questions for validation, and a private test set. Since the testset is private, for this dataset, we report all the result on validation set.
ST-VQA contains 23K images and 31K questions, with 16K images and 22K questions for training, and 2.8K images with 4K questions for testing.
ChartQA includes 21.6K charts with 32.3K question-answer pairs, split into 19K charts with 28K questions for training, 1K charts with 1.8K questions for validation, and 1.6K charts with 2.5K questions for testing.
OKVQA consists of 14K images with 14K questions, divided into 9K questions for training, 5K for testing.%, where each question has a ground-truth answer along with multiple human-annotated acceptable answers.

\begin{table*}[t!]
\centering
\small
\begin{tabular}{llcccc}
\toprule
\textbf{Task} & \textbf{Dataset} & \textbf{Metric} & \textbf{S-VLM (Zero-shot)} & \textbf{S-VLM (HL)} & \textbf{S-VLM (MPA)} \\
\midrule
OCR & ICDAR2015 & WRR & 31.9 & 33.2 & \textbf{36.4} ($\uparrow 4.5$) \\
\midrule
\multirow{3}{*}{TC} & \multirow{3}{*}{TextCaps} & BLEU-1 & 7.9 & 13.4 & \textbf{15.3} ($\uparrow 7.4$) \\
 &  & ROUGE-L & 17.4 & 18.3 & \textbf{20.6} ($\uparrow 3.2$) \\
 &  & CIDEr   & 8.7  & 34.6 & \textbf{38.1} ($\uparrow 29.4$) \\
\bottomrule
\end{tabular}
\caption{Comparison of OCR and text-aware captioning performance. Despite using no ground-truth labels, \textsc{MPA} outperforms both the zero-shot baseline and models trained on human-labeled data (HL).}
\label{tab:ocr_tc_appendix}
\end{table*}

\section{Implementation Details}
\label{sec:appendix}
We implement our method using PyTorch. Majority of the chosen \svlm{}s and \lvlm{}s employed in our propsed method MPA, we use their original code-base repositories and/or their Huggingface implementations depending on the ease of reproducibility. Parity leveler (Section 3.3) module trains the entire \svlm{} on the samples obtained from the PI module (Section 3.2) for one epoch, for all the benchmark datasets. Hyperparameters used by the PL module for different \svlm{}s are summarized in Table~\ref{tab:batch_lr}. All our experiments are conducted on a machine with three Nvidia A6000 GPUs (48 GB each). For every \lvlm{} and \svlm{} combination, it took approximately, 5-12 GPU hours for entire MPA, for one dataset. We use \texttt{gpt-4o-2024-11-20}~\cite{openai2024gpt4o} for our closed-source \lvlm{} ablation. 

\begin{table}[t!]
  \centering
  \resizebox{0.6\columnwidth}{!}
  {
  \begin{tabular}{l c c}
  \toprule
    \svlm{} & Batch Size & LR \\
    \midrule
    \multirow{1}{*}{Qwen2VL-2B} & 16 & 1e-5 \\
    \multirow{1}{*}{InternVL2-2B} & 16 & 4e-5 \\
    \multirow{1}{*}{InternVL2-4B} & 6 & 4e-5 \\
    \multirow{1}{*}{SmolVLM-500M} & 16 & 1e-4 \\
    \multirow{1}{*}{TinyLLaVA-2B} & 16 & 1e-4 \\
    \bottomrule
    \end{tabular}
  }
  \caption{Hyperparameters used in the parity leveler module (Section 3.3) for each \svlm{}.}
  \label{tab:batch_lr}
\end{table}

\section{Prompts used}
In this section, we provide the VLM prompts used in the PA module (Section~\ref{subsec:pa}) to generate pseudo-annotations for all four datasets: 

% \noindent (i) Prompt for TextVQA and STVQA: 
\begin{tcolorbox}[title=PA prompt for: TextVQA and ST-VQA, colback=yellow!10, colframe=brown!50, coltitle=black]
\footnotesize
\textcolor{blue}{$<$image$(\mathbf{I})$$>$} \\
The objective is to generate a question-answer pair for a Textual Visual Question Answering (Text-VQA) task. Your task is to create a contextually relevant question that directly relates to the image's content, incorporating reasoning or direct references to the text, and its correct answer. \\
Output: \\
- Question: A natural language question grounded in the image's content and text. \\
- Answer: A concise response (single word, phrase, or Yes/No) derived from the text or reasoning based on it. \\
Assistant: \textcolor{red}{Question: $\tilde{\mathbf{Q}}$, Answer: $\tilde{\mathbf{A}}$}
% \label{prompt_textvqa}
\end{tcolorbox}

% \noindent (ii) ChartQA~\cite{masry2022chartqa}:

\begin{tcolorbox}[title=PA prompt for: ChartQA, colback=yellow!10, colframe=brown!50, coltitle=black]
\footnotesize
\textcolor{blue}{$<$chart image $(\mathbf{I})$$>$} \\
The objective is to generate a question-answer pair for a Chart Visual Question Answering (ChartVQA) task. Your task is to create a contextually relevant question that directly relates to the content of a given chart, incorporating reasoning based on the visualized data. \\
Output Requirements: \\
- Question: A natural language question grounded in the chart's content, requiring numerical reasoning, trend analysis, or data lookup. \\
- Answer: A concise response (single word, number, phrase, or Yes/No) derived from the chart’s data. \\
Guidelines for Question Generation: \\
1. Direct Lookup Questions – extracting specific values from the chart. \\
2. Comparison Questions – comparing values between different categories. \\
3. Trend \& Pattern Recognition – identifying increases, decreases, or correlations in the data. \\
4. Inference-Based Questions – requiring reasoning beyond direct value lookup. \\
Ensure the question is meaningful and the answer is accurate based on the chart data. \\
Assistant: \textcolor{red}{Question: $\tilde{\mathbf{Q}}$, Answer: $\tilde{\mathbf{A}}$}
% \label{prompt_chartvqa}
\end{tcolorbox}

% \noindent (iii) OKVQA~\cite{okvqa}:

\begin{tcolorbox}[title=PA prompt for: OKVQA, colback=yellow!10, colframe=brown!50, coltitle=black]
\footnotesize
\textcolor{blue}{$<$image$(\mathbf{I})$$>$} \\
The objective is to generate a question-answer pair for a Knowledge-based Visual Question Answering (K-VQA) task. Your task is to create a contextually relevant question that directly relates to the image's content while requiring external world knowledge to answer correctly, and its correct answer. \\
Output Requirements: \\
- Question: A natural language question grounded in the image’s content but requiring reasoning beyond direct perception, incorporating real-world knowledge. \\
- Answer: A single-word response based on general world knowledge. \\
Guidelines for Question Generation: \\
1. Object \& Scene Understanding – identifying objects or actions in the image and connecting them to broader knowledge. \\
2. Commonsense Reasoning – requiring logical deductions about the scene. \\
3. Cultural \& Historical Context – related to well-known historical events, traditions, or cultural references. \\
4. Scientific \& Factual Knowledge – involving basic physics, biology, geography, or general knowledge. \\
5. Everyday Life \& Social Understanding – questions about daily activities, professions, or human behaviors. \\
\# Ensure that the generated question is meaningful and requires external knowledge beyond just the image’s visual content. \\
Assistant: \textcolor{red}{Question: $\tilde{\mathbf{Q}}$, Answer: $\tilde{\mathbf{A}}$}
% \label{prompt_kvqa}
\end{tcolorbox}

Note that, to ensure fair comparison, the pseudo-annotation prompts are same for all variants of \textsc{L-VLM}s used. Further, the prompt we used for QA is `\texttt{Answer the following question in a single word or phrase}', which is common for all datasets across all \textsc{S-VLM}s.

\begin{figure*}[!t]
\centering
  \includegraphics[width=\columnwidth]{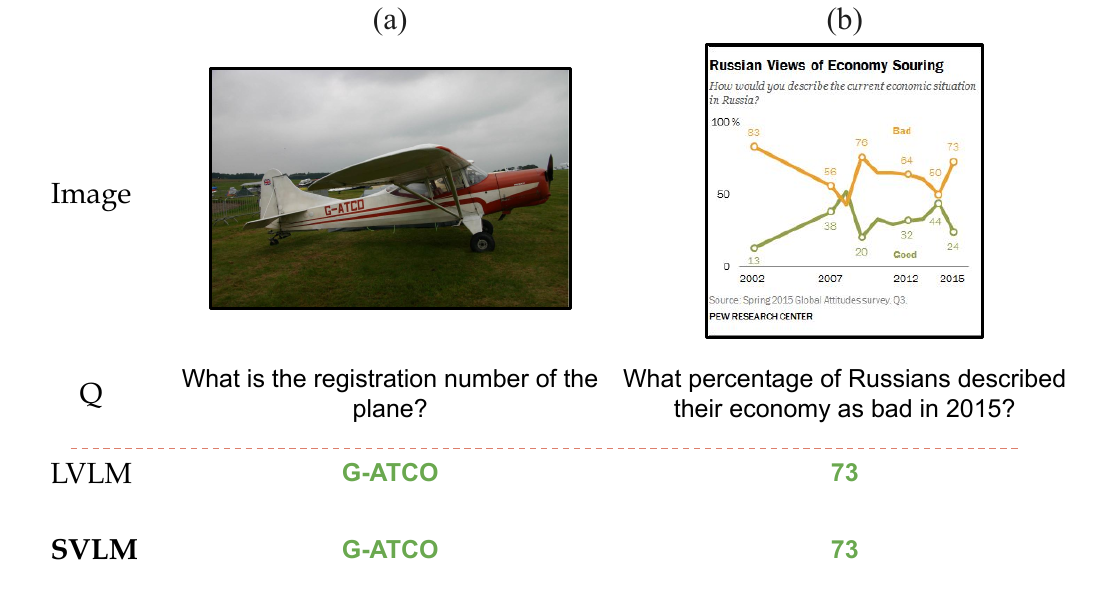}
  \includegraphics[width=\columnwidth]{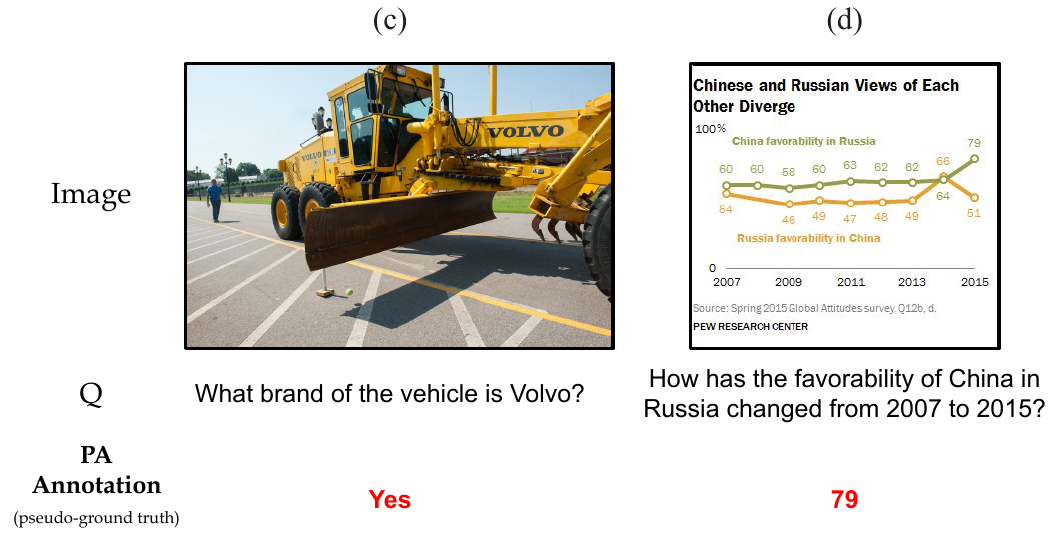}
  \caption{\textbf{Left two examples:} Pseudo-annotations discarded by PI module as they do not constitute knowledge-gap. \textbf{Right two examples:} Pseudo-annotations discarded by PI module as they are noisy annotations.}
  \label{fig:notgap}
\end{figure*}

\section{Qualitative Results}
\begin{figure*}[!t]
\centering
  \includegraphics[width=\textwidth]{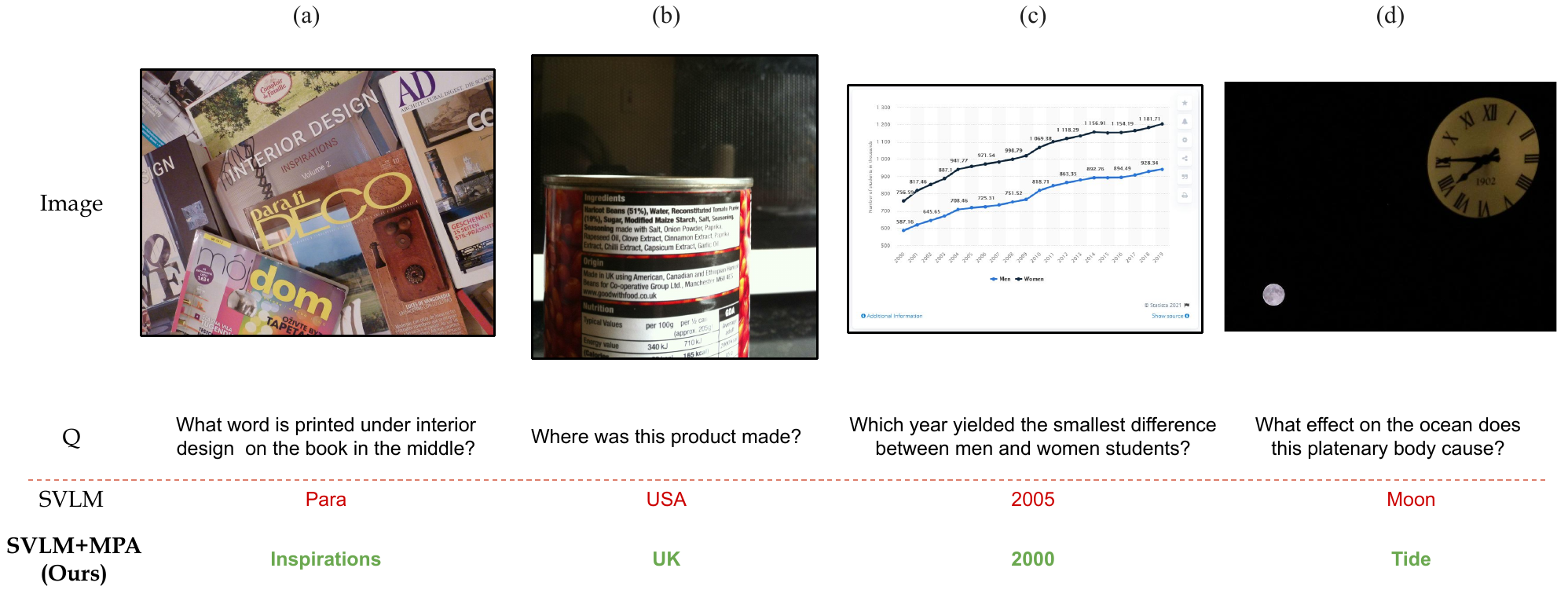}
  \caption{A selection of results showing zero-shot SVLM versus MPA-aligned SVLM. MPA config: \svlm{}: Qwen2VL-2B, \lvlm{}: Qwen2VL-7B. Green and red text correspond to correct and incorrect answers, respectively. \textbf{(Best viewed in color)}}
  \label{fig:visRes2}
\end{figure*}

\begin{figure*}[!t]
\centering
  \includegraphics[width=\textwidth]{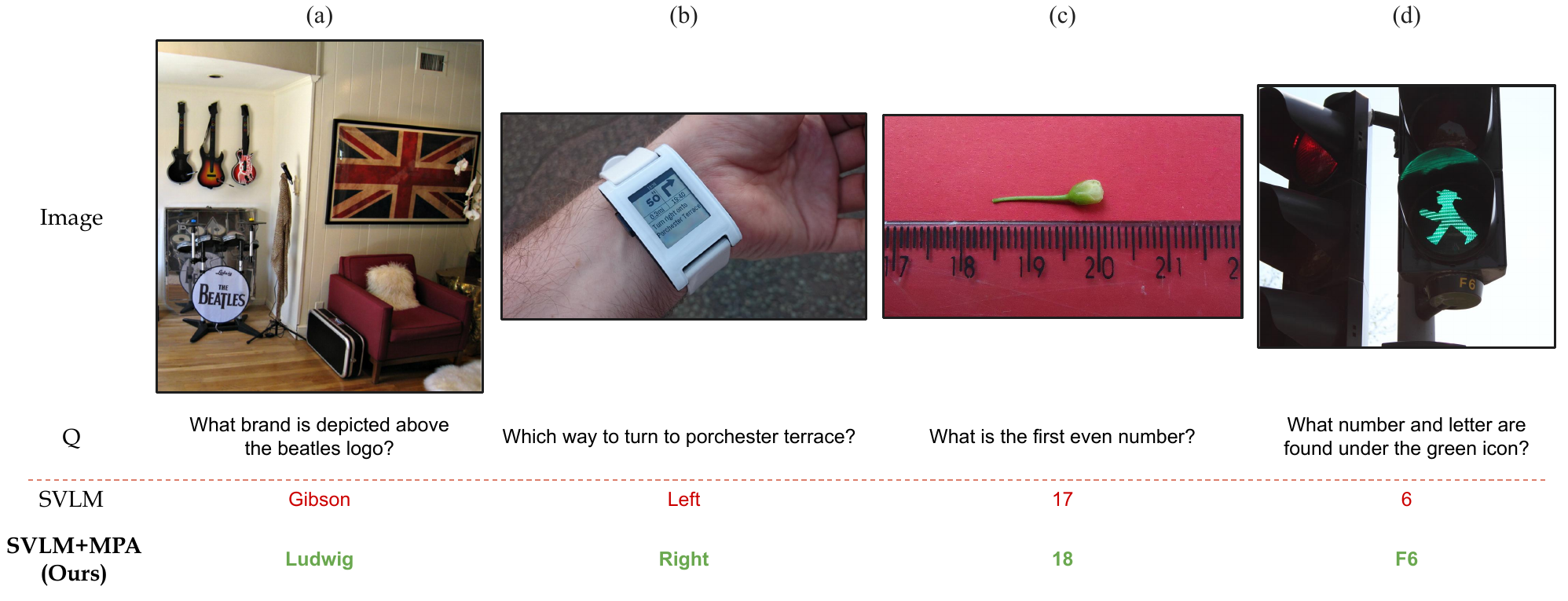}
  \caption{Few more results from TextVQA showing the efficiency of MPA-aligned \svlm{} over baseline \svlm{}.}
  \label{fig:textvqa_addn_results}
\end{figure*}

\begin{figure*}[!t]
\centering
  \includegraphics[width=\textwidth]{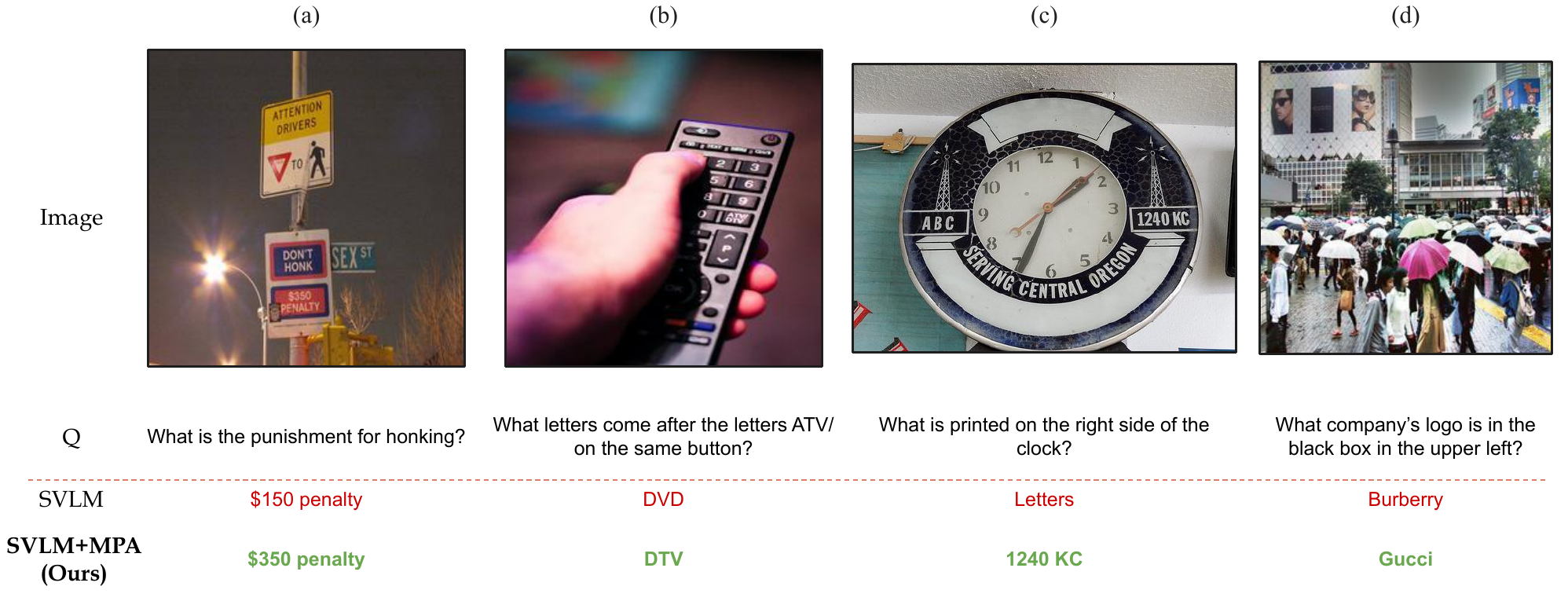}
  \caption{Few more results from STVQA showing the efficiency of MPA-aligned \svlm{} over baseline \svlm{}. Green and red text correspond to correct and incorrect answers, respectively. \textbf{(Best viewed in color)}}
  \label{fig:stvqa_addn_results}
\end{figure*}

\begin{figure*}[!t]
\centering
  \includegraphics[width=\textwidth]{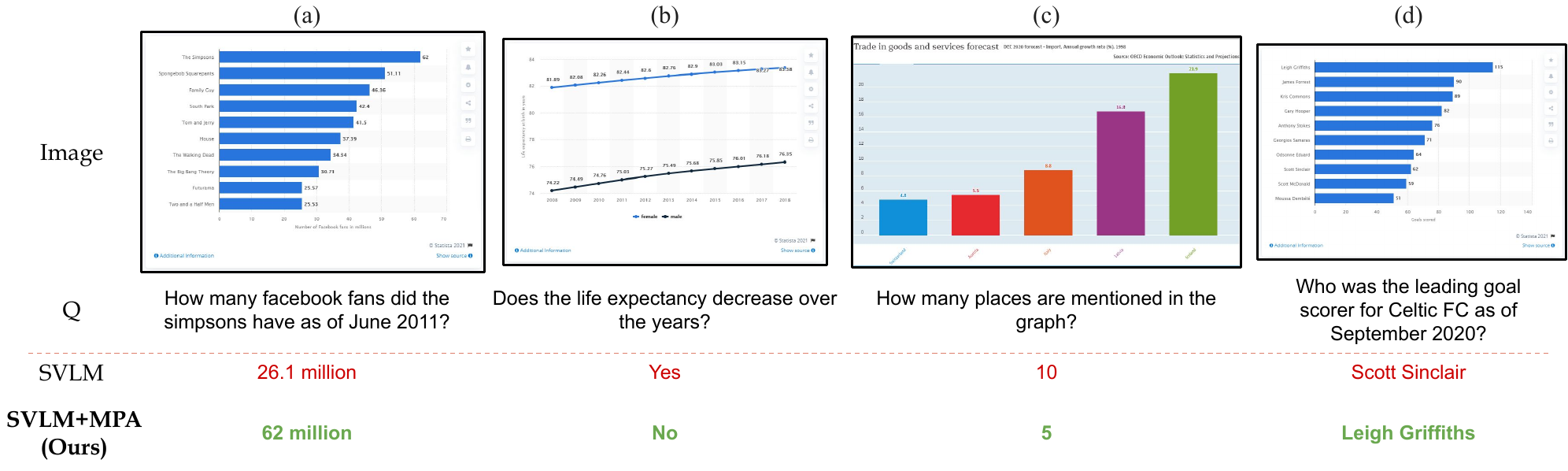}
  \caption{Few more results from ChartQA showing the efficiency of MPA-aligned \svlm{} over baseline \svlm{}. Green and red text correspond to correct and incorrect answers, respectively. \textbf{(Best viewed in color)}}
  \label{fig:chartvqa_addn_results}
\end{figure*}

\begin{figure*}[!t]
\centering
  \includegraphics[width=\textwidth]{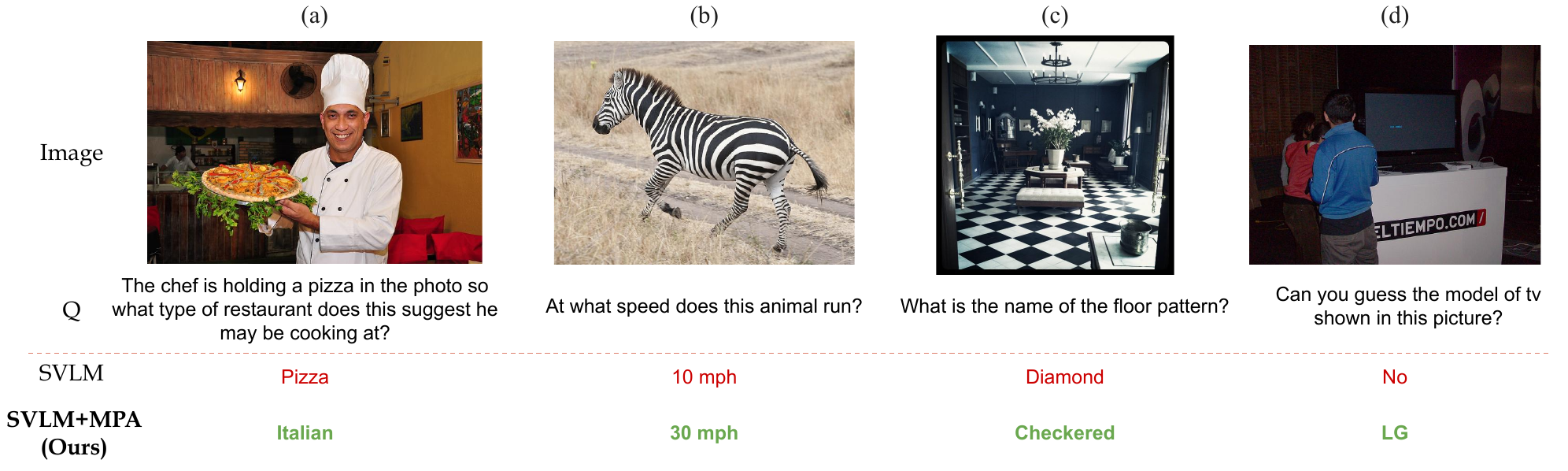}
  \caption{Few more results from OKVQA showing the efficiency of MPA-aligned \svlm{} over baseline \svlm{}. Green and red text correspond to correct and incorrect answers, respectively. \textbf{(Best viewed in color)}}
  \label{fig:aokvqa_addn_results}
\end{figure*}

Figure~\ref{fig:visRes2} presents a selection of examples where MPA alignment enables S-VLM to correct errors made by the original zero-shot S-VLM. From a rigorous examination of the results, we find that MPA significantly improves performance in visual text reasoning, plot interpretation, and knowledge-based question answering. Further, we show additional qualitative samples for showing zero-shot SVLM versus MPA-aligned \svlm{} across all four datasets: TextVQA in Figure~\ref{fig:textvqa_addn_results}, ST-VQA in Figure~\ref{fig:stvqa_addn_results}, ChartQA in Figure~\ref{fig:chartvqa_addn_results} and OKVQA in Figure~\ref{fig:aokvqa_addn_results}.

Furthermore, in Figure~\ref{fig:notgap}, we show selected examples that do not represent a disparity between \svlm{} and \lvlm{} ((a), (b)), and another set of examples that are noisy annotations ((c), (d)), both of which are discarded by the PI module.

\end{document}